\documentclass[10pt,twocolumn,letterpaper]{article}

\usepackage[pagenumbers]{cvpr} 

\definecolor{cvprblue}{rgb}{0.21,0.49,0.74}

\usepackage{bbm}
\usepackage{multirow,makecell}
\usepackage{xcolor}
\definecolor{myforestgreen}{RGB}{34, 139, 34}  
\usepackage{ragged2e} 

\usepackage{colortbl}
\usepackage{tcolorbox}

\usepackage[pagebackref,breaklinks,colorlinks,allcolors=cvprblue]{hyperref}

\title{Video2Layout: Recall and Reconstruct Metric-Grounded Cognitive Map for Spatial Reasoning}

\author{
  Yibin Huang$^{1}$,
  ~Wang Xu$^{2}$\thanks{Corresponding Author.},
  ~Wanyue Zhang$^{3}$, 
  ~Helu Zhi$^{1}$,
  ~Jingjing Huang$^{2}$, \\
  Yangbin Xu$^{4}$,
  ~Yangang Sun$^{2}$,
  ~Conghui Zhu$^{1}$\footnotemark[1], 
  ~Tiejun Zhao$^{1}$ \\
  $^{1}$ Faculty of Computing, Harbin Institute of Technology ~$^{2}$ Tsinghua University \\
  $^{3}$ Institute of Automation, Chinese Academy of Sciences \\
  $^{4}$ Institute of Microelectronics of the Chinese Academy of Sciences \\
  {\tt\small yibinhuang@stu.hit.edu.cn, xwjim812@126.com}
}

\begin{document}
\maketitle

\begin{abstract}

Spatial intelligence is a critical frontier for Multimodal Large Language Models (MLLMs), empowering them to comprehend the physical world. 
Drawing inspiration from human perception mechanisms, prior studies attempt to construct a spatial understanding via grid-based cognitive maps.
However, current grid-based map methods rely on discretized representations, which limit the model's ability in fine-grained spatial reasoning. 
To overcome this limitation, we propose Video2Layout, a framework for reconstructing metric-grounded spatial layouts from video. 
The framework uses continuous object boundary coordinates to enable quantitative spatial computation, which effectively reduces ambiguity in natural language descriptions of spatial relationships.
Specifically, our method comprises two stages. First, in supervised fine-tuning stage, we construct a high-quality dataset from the AI2THOR simulator, which enables the model to learn the mapping from visual inputs to precise boundary coordinates. Subsequently, a reinforcement fine-tuning stage enhances the model’s real-world generalization capabilities.
Based on the above framework, we investigate factors that affect cognitive map accuracy and quantify its relationship with task performance. 
Evaluated on mainstream spatial reasoning benchmarks, our model, V2LO-7B achieves an average improvement of 3.24\% over the model trained on grid maps, validating the superiority of our method.

\end{abstract}    
\begin{figure}[t!]
    \centering
    \includegraphics[
        width=\linewidth,
        height=0.5\textheight,  
        keepaspectratio,
        trim=1.5cm 2.5cm 1.5cm 3.0cm, 
        clip
    ]{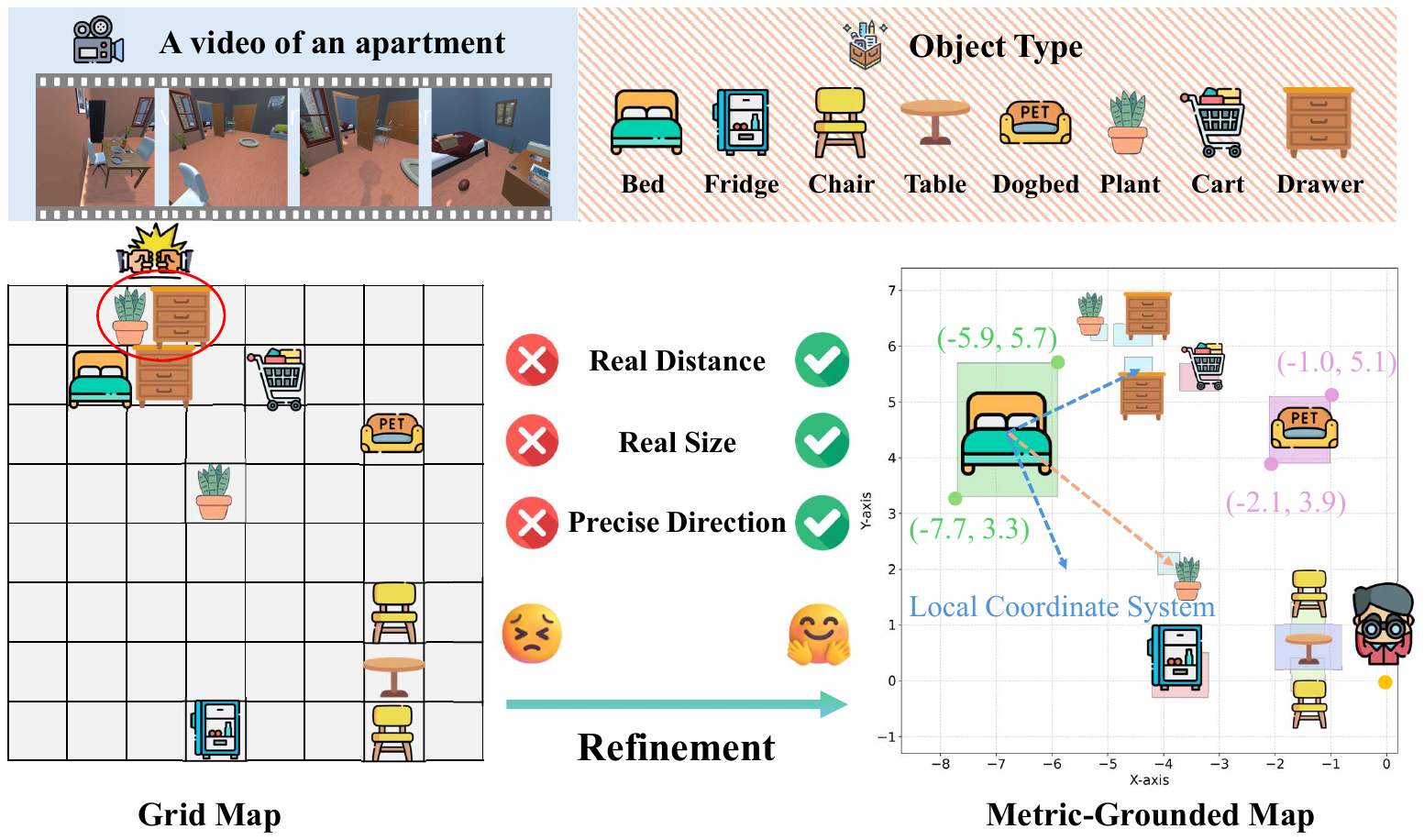}
    \caption{Comparison of cognitive map representations. A conventional grid map (left) introduces metric and semantic inaccuracies regarding real distance, object size, and precise direction. In contrast, our method generates a metric-grounded map (right) that assigns precise Bird’s-Eye View (BEV) coordinates to objects in an observer-centered perspective, establishing a quantitative foundation for fine-grained spatial reasoning.}
    \label{fig:show}
\end{figure}

\section{Introduction}
\label{sec:intro}

With the rapid advancement of multimodal large language models~\cite{li2024llava, Qwen2.5-VL, chen2024expanding}, spatial intelligence~\cite{wang2024robogen} has become a key research frontier, serving as a core link toward embodied intelligence~\cite{chiang2024mobility, cai2025cookbench}. Building spatial understanding capabilities requires models to possess refined perception, robust imagination, and rigorous reasoning~\cite{yin2025spatial}.  In practice, current MLLMs still demonstrate shortcomings in spatial perception and reasoning~\cite{zhang2025mllms, zhao2025embodied}. The core bottleneck lies in the difficulty of effectively aggregating multi-frame local spatial features into a coherent and unified global spatial representation~\cite{yang2025thinking}.

Drawing from cognitive neuroscience~\cite{nadel1991hippocampus, ishikawa2021spatial}, the core of humans' efficient mastery of spatial reasoning capabilities lies in the sophisticated cognitive map mechanism. Building on this principle, prior studies~\cite{manh2025mind, su2025reactive, de2024learning} propose that models can simulate human cognitive logic by explicitly constructing a cognitive map. However, existing map construction approaches~\cite{yang2025thinking, yin2025spatial, ouyang2025spacer} construct M×M grid-based cognitive map, which discretizes continuous space. These maps describe inter-object spatial relationships in a coarse-grained manner, and only approximate real-world spatial metrics, and are prone to object overlap within a single grid, which can be suboptimal for fine-grained spatial reasoning tasks.

To address these intrinsic limitations, we propose an innovative Video2Layout framework shown in \cref{fig:show}. Unlike grid maps, this framework generates a metric-grounded cognitive map composed of BEV bounding boxes to achieve supersensing~\cite{yang2025cambrian}.
This map, defined by continuous, real-world coordinates, enables the model to learn and construct a physically-accurate spatial representation from video. Such a metric-aware map provides a solid foundation for rigorous spatial computation~\cite{wularge}, which mitigates the inherent ambiguity in natural language-driven reasoning.
To enable the model to generate such a map in real scenes, our training paradigm unfolds in two stages. First, in the Supervised Fine-Tuning (SFT) stage, we constructed a synthetic dataset using the AI2THOR simulator~\cite{ai2thor}. 
This dataset aims to address key challenges in real-world spatial data, including high acquisition costs, notable data noise, and limited accuracy in target localization.
We leverage the precise coordinate information from simulator as a supervisory signal to ground the model's spatial perception and computation~\cite{ma2025spatialreasoner}. 
In the subsequent Reinforcement Fine-Tuning (RFT) stage, to alleviate the gap between simulated data and real scenarios, we adopt the GRPO algorithm~\cite{shao2024deepseekmath} for policy optimization, which enhances the model’s generalization performance. 
Through two stages of training, the spatial reasoning ability of the model has been effectively improved.

To conduct a comprehensive study of cognitive maps, we conduct in-depth investigations into how various factors influence map accuracy and quantify the relationship between map accuracy and task scores. We further comprehensively evaluate mainstream spatial reasoning benchmarks, where our V2LO-7B model achieves an average improvement of 3.24\% over grid-map-based counterparts, thus validating the effectiveness of our method.

In summary, our key contributions are as follows:
\begin{itemize}
    \item We propose Video2Layout, an innovative framework that integrates metric-grounded cognitive map and SFT-to-RL training paradigm to enhance models’ spatial reasoning capabilities in real-world scenarios.
    
    \item We conduct in-depth analysis into how diverse factors influence map accuracy and quantify the relationship between map accuracy and task scores.
    
    \item Through extensive experiments, our V2LO-7B model achieves a performance gain of 3.24\% on average over grid-map-based baseline, validating the effectiveness of our method.
\end{itemize}
\section{Related Work}
\label{sec:related_word}

\begin{figure*}[t!]
    \centering
    \includegraphics[width=\linewidth, trim=0cm 6.5cm 0cm 6.5cm, clip]{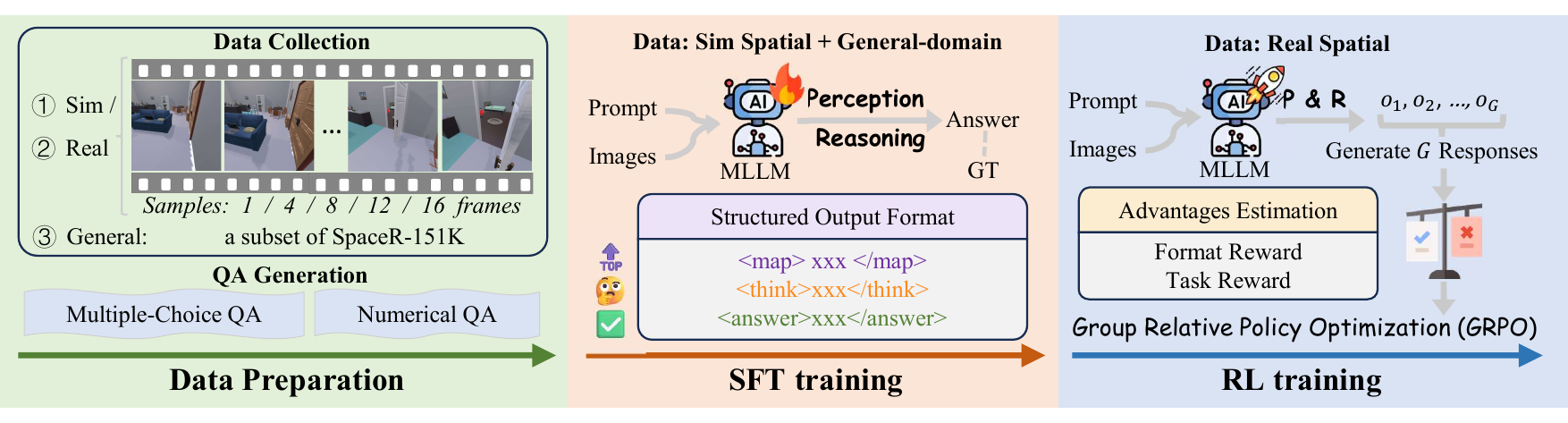}
    \caption{The overall framework diagram of Video2Layout. (1) Data preparation stage focuses on generating QA pairs from simulated spatial, real spatial, and general-domain data sources. (2) SFT stage aims to train the model on simulated spatial data and general-domain data, enabling model to generate a metric-grounded map and adopt a structured reasoning output format. (3) RFT stage leverages the GRPO algorithm for training on real-world spatial data, effectively realizing the generalization of real scenarios.}  
    \label{fig:pipeline}
\end{figure*}

Spatial reasoning is a pivotal research area within MLLMs, where the current development paradigm is advancing along four key dimensions: 1) constructing high-quality, meticulously annotated datasets that cover diverse spatial scenarios to support model training and evaluation~\cite{zhang2025flatland, chen2024spatialvlm, xu2025multi}; 2) innovating model architectures by introducing specialized modules, including 3D point cloud encoders~\cite{zheng2025video, wang2025spatialclip, qi2025gpt4scene} and depth encoders~\cite{cai2025spatialbot, cheng2024spatialrgpt, liu2025ssr}, to enhance cross-modal feature alignment and fusion; 3) developing efficient task-specific training methodologies, such as SFT and RL, to refine model performance on specific spatial reasoning tasks~\cite{wu2025st, zhou2025roborefer, wang2025svqa}; and 4) employing customized reasoning strategies to guide the model's spatial reasoning process. Synergistic progress across these four domains enhances the model's proficiency in reasoning within complex spatial tasks.

Among the four dimensions, the fourth one—employing customized reasoning strategies—has become an active research frontier, where the CoT reasoning process serves as a core implementation. While general free-text CoT has achieved remarkable success in general-purpose reasoning tasks, it exhibits sub-optimal performance in the spatial domain, primarily due to its ambiguous free-form narratives that fail to support tasks requiring accurate geometric localization and perspective transformation~\cite{zheng2025multimodal, yang2025thinking}. This limitation has spurred a paradigm shift toward structured CoT, which incorporates explicit geometric and relational constraints (e.g., coordinates and reference frames). This structured approach has yielded significant performance improvements in challenging spatial tasks such as complex perspective-taking~\cite{li2024topviewrs, omnispatial25, wu2025scog}, with representative works including SpatialPIN~\cite{ma2024spatialpin}, SpatialPrompt~\cite{liao2024reasoningpathsreferenceobjects}, and SpatialMind~\cite{zhang2024spatial}.

The ``cognitive map'' approach, an important branch of customized spatial reasoning that leverages coordinate information~\cite{manh2025mind, su2025reactive, ishikawa2021spatial}, can enhance the model’s spatial reasoning capabilities but faces three key challenges in current implementations. First, most existing studies adopt a discrete M×M grid map, which has limited capacity to accurately capture the actual distance between objects and their true sizes~\cite{yin2025spatial, ouyang2025spacer}. Second, some studies rely on external modules such as depth estimation components, which increases the overall model complexity ~\cite{gholami2025spatial, zhang2024visual, ivanov2025mapfm}. Third, some works are limited to single-frame image inputs rather than sequential video inputs, failing to meet the requirements of dynamic spatial reasoning scenarios~\cite{ma2025spatialreasoner, chang2025vlm, huang20253d}.

The above studies indicate that directly generating fine-grained object coordinates from video remains insufficiently explored. Consequently, we propose Video2Layout, a framework that reconstructs metric-grounded spatial layouts and utilizes structured CoT to enhance models' spatial reasoning.
\section{Methodology}
\label{sec:method}

The overall framework comprises three sequential components, as illustrated in \cref{fig:pipeline}. First, it starts with data preparation, which involves curating a comprehensive dataset encompassing simulated and real-world spatial data, as well as general VQA data, with detailed specifications elaborated in \cref{sec:dataset_benchmark}. This is followed by the supervised fine-tuning (SFT) phase (detailed in \cref{sec:sft}), where the model is fine-tuned on simulated spatial data and general VQA data to generate a metric-grounded map and establish a Structured CoT reasoning process. Finally, in the reinforcement fine-tuning (RFT) phase (presented in \cref{sec:rl}), we adopt the GRPO algorithm to perform policy optimization on real-world spatial data, thereby effectively bridging the sim-to-real gap.

\begin{figure*}[t!]
    \centering
    \includegraphics[width=\linewidth, trim=1.0cm 7cm 1.0cm 6.5cm, clip]{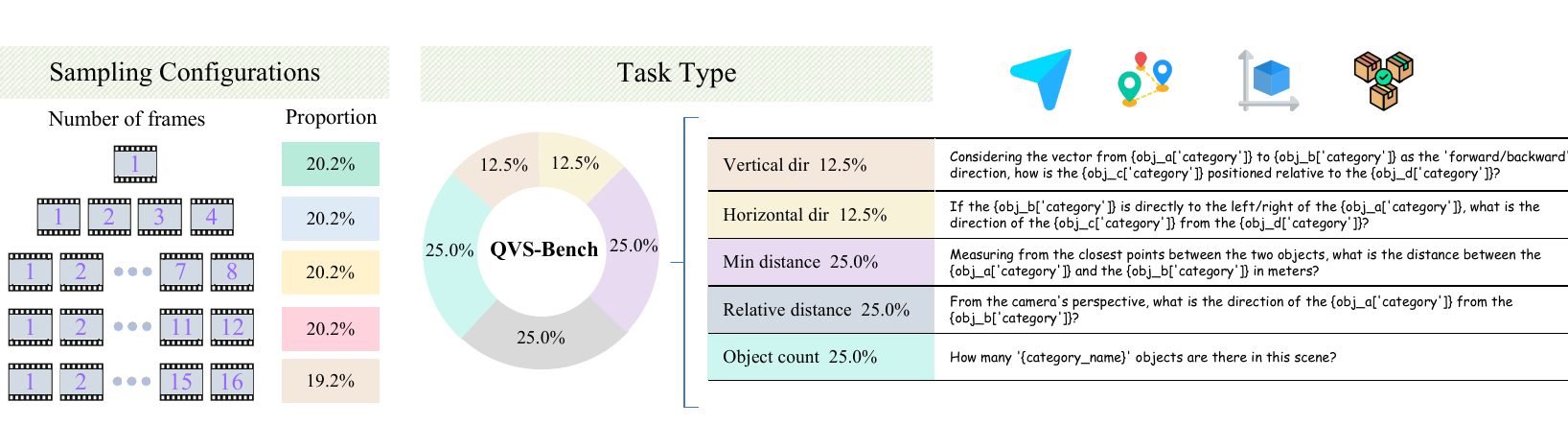}  
    \caption{A detailed overview of the QVS-Bench benchmark. Left: The ``Sampling Configurations'' panel depicts the proportional composition of data samples across five distinct input frame counts (1, 4, 8, 12, and 16). Right: The ``Task Type'' panel details the proportional breakdown of the five spatial reasoning tasks, with each category accompanied by its corresponding example question template.}
    \label{fig:benchmark}  
\end{figure*}

\subsection{Dataset and Benchmark}
\label{sec:dataset_benchmark}

\textbf{Overview.}
To enhance the model's perceptual capabilities for metric-grounded maps, we present the V2LO-28K dataset, which is partitioned into three subsets.

\begin{itemize}
    \item The SFT training set (16K samples) consists of 12K samples of simulated spatial data and 4K samples of general VQA data. Specifically, the 12K simulated spatial data enable the model to learn the mapping relationship from visual inputs to precise coordinates, while the 4K general scene data samples ensure the retention of the model’s general domain capabilities. 
    
    \item The RL training set (8K samples), derived from the ScanNet dataset, is specifically designed to enhance the model’s adaptability to real-world scenarios and further improve its spatial reasoning performance. 
    
    \item The QVS-Bench (4K samples), also derived from ScanNet, is strictly partitioned by scene ID and completely isolated from the RL training set to avoid data leakage. This benchmark serves as a dedicated testbed to verify whether our RL approach achieves effective sim-to-real generalization.
\end{itemize}

\textbf{Data Collection.}
To generate our spatial reasoning data, we employ a unified pipeline to process both simulated and real scenes.

\begin{itemize}
    \item For simulated data, a pre-defined path planning algorithm is first employed to generate motion trajectories, along which cameras capture videos. These videos are then segmented and filtered using object-centric metadata (e.g., bounding box size and pixel coverage area) to isolate clips containing salient targets. From these validated clips, continuous image sequences with lengths of 1, 4, 8, 12, and 16 frames are systematically extracted.
    
    \item For real data, a consistent workflow is adopted on the ScanNet dataset: scene metadata are parsed, RGB frames are resampled to 20 FPS to ensure temporal consistency, and the identical multi-scale sampling strategy is subsequently applied to extract image sequences of the same lengths (1, 4, 8, 12, and 16 frames). 
    
    \item For general VQA data, we randomly sample 4K samples from the Video-R1-260k dataset to construct a general corpus.
\end{itemize}

\textbf{QA Generation.}
Leveraging the parsed meta-information, we propose an automatic QA pair generation framework tailored to spatial reasoning tasks. 
As depicted in \cref{fig:benchmark}, the generated QA pairs are systematically categorized into two distinct types: 1) Multiple-choice QA, which focuses on spatial relationship judgment and encompasses three tasks: relative distance, vertical directional relationships, and horizontal directional relationships; 2) Numerical QA, which focuses on spatial quantity calculation and covers two tasks: minimum distance, and object count.

These generated QA pairs form the basis for training data in the subsequent SFT and RFT phases.
For the SFT stage, structured CoT reasoning processes will be added as part of the answer, which is generated via predefined templates and refined by Qwen3-30B-A3B-Instruct. The structured CoT processes provide supervision for coordinate representation and numerical calculation learning. 
The coordinate system is selected based on the starting position of the first frame and the initial orientation of the camera.
For the RL stage, only QA pairs are retained without CoT processes, adopting a result supervision approach that dispenses with fine-grained annotation of real-world data. 

\subsection{Supervised Finetune}
\label{sec:sft}

\begin{figure*}[t!]
    \centering
    \includegraphics[width=\linewidth, trim=0cm 5cm 0cm 5cm, clip]{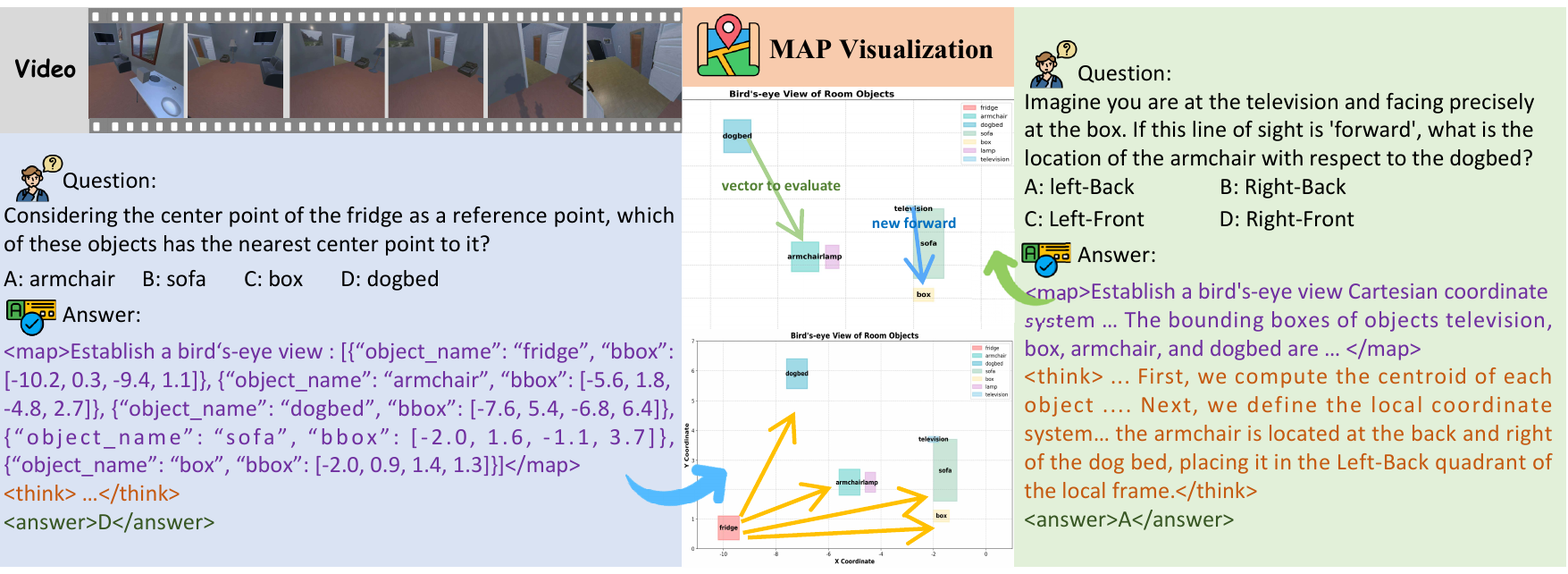}
    \caption{Illustrative examples of the structured spatial reasoning process. The structured CoT follows a unified approach: mapping objects onto a metric-grounded map to convert spatial reasoning into mathematical computation.}
    \label{fig:dataset_example}
\end{figure*}

As a core component of the Video2Layout framework, this phase is designed to guide the model to generate a metric-grounded map while upholding a structured reasoning output format. 
To achieve this, we introduce a schema consisting of distinct functional modules, as illustrated in \cref{fig:dataset_example}. 
Our schema explicitly decouples spatial perception from subsequent logical deliberation. 
Specifically, we employ a structured chain-of-thought process with three key modules, which work synergistically to guarantee accurate spatial reasoning.

\begin{itemize}

    \item \textbf{Map Module:} This module is responsible for spatial perception and formal scene representation. It constructs a structured bird’s-eye view within a Cartesian coordinate system, which is defined according to the starting position and initial camera orientation of the first frame. By projecting relevant objects to their corresponding bounding box coordinates, the module effectively alleviates the ambiguities inherent in natural language descriptions of object locations, thus improving the accuracy of spatial localization.
    
    \item \textbf{Think Module:} This module undertakes the deductive reasoning stage. Built upon the coordinate-based representation provided by the Map Module, our method performs explicit, step-by-step mathematical and logical operations. For example, to answer minimum distance queries, it computes the Euclidean distance between specified objects according to their bounding box coordinates. For more complex relational reasoning tasks—such as determining an object’s position from a novel viewpoint—this module dynamically establishes a local coordinate system, projects target object locations into this system, and then applies vector operations (e.g., dot products or cross products) to determine their precise orientation. Such a numerical computation-based approach effectively addresses the inherent inaccuracy of spatial reasoning that relies solely on natural language.
    
    \item \textbf{Answer Module:} This module generates the final output directly from the explicit reasoning process of the Think Module. It transforms structured reasoning results, such as computed distances and orientation judgments, into task-aligned formats including natural language answers and standardized numerical outputs.
    
\end{itemize}

Notably, the Map and Think modules are optional and task-dependent. Specifically, if a given question involves no spatial queries (e.g., general non-spatial VQA questions) or does not require complex deductive reasoning, the corresponding module can be omitted. In such scenarios, the framework directly produces the final output through the Answer Module, yielding strong efficiency and flexibility. Overall, by enabling the synergistic coordination of the three modules, the SFT phase establishes a robust foundation for the subsequent RFT phase and the overall spatial reasoning capability of the model.

\subsection{Reinforcement Finetune}
\label{sec:rl}

The training data utilized in the preceding SFT phase is predominantly simulated, which inherently incurs distributional discrepancies relative to real-world spatial scenarios. Targeting this critical flaw, in the RFT stage, we leverage the GRPO algorithm to train the model on real-world spatial data, thereby effectively transferring and extending the spatial representation capabilities to real-world scenarios and bridging the sim-to-real gap.

Considering that 1) most spatial reasoning tasks only concern the relative positional relationships between objects, and 2) continuous camera motion during navigation tends to induce unstable reward calculation, we adopt direct result-level supervision. Based on the GRPO algorithm, we design two verifiable reward functions—Format Reward and Task Answer Reward, which jointly guide the policy optimization of the model.

\textbf{Format Reward.} To ensure the model responses adhere to a predefined structure, we define a format reward $R_{\text{format}}$:
\begin{equation} \label{eq:fmt_reward}
R_{format}(\hat{y}) = 
\begin{cases} 
1, & \text{if $\hat{y}$ matches format}, \\
0, & \text{otherwise}.
\end{cases}
\end{equation}

\textbf{Multi-choice Reward.} For multi-choice QA, the reward $R_{\text{mc}}$ is binary, based on exact match with the ground truth: 
\begin{equation} \label{eq:multi-choice_reward}
R_{mc}(\hat{y}, y) = 
\begin{cases} 
1, & \text{if } \hat{y} = y, \\
0, & \text{otherwise},
\end{cases}
\end{equation}
where $\hat{y}$ is the model's response and $y$ is the ground truth.

\textbf{Numerical Reward.} To assess numerical values, we compute relative accuracy across varying confidence thresholds $\theta_i \in \{0.5, 0.55, \dots, 0.95\}$. The numerical reward $R_{num}$ is defined as: 
\begin{equation} \label{eq:numerical_reward}
R_{num}(\hat{y}, y) = \frac{1}{N} \sum_{i=1}^{N} \mathbbm{1} \left( \frac{\left|\hat{y} - y\right|}{y} \leq 1 - \theta_i \right),
\end{equation}
$N$ is the number of confidence thresholds.

Formally, the reward $R_i$ is defined as:
\begin{equation}\label{eq:map_aug_reward}
R_i = \alpha*R_{format} + (1-\alpha)*R_{task} 
\end{equation}
where $task \in \{mc, num \}$ and $\alpha=0.1$.
The advantage $A_i$, representing the relative quality of the $i$-th response $o_i$, is computed over the updated rewards within each group of responses $[o_1, o_2, \cdots, o_G]$, where $G$ is the number of output responses. 

\begin{equation}\label{eq:advantage}
A_i = \frac{R_i - \operatorname{mean}(\{R_j\}_{j=1}^G)}{\operatorname{std}(\{R_j\}_{j=1}^G)}.
\end{equation}

The final policy update follows the clipped surrogate objective of GRPO.

\begin{equation}
\begin{split}
J(\theta) = \mathbb{E}_{q,\{o_i\}} 
\Bigg[ \frac{1}{G} \sum_{i=1}^{G} \min\Bigg( & \frac{\pi_\theta(o_i|q)}{\pi_{\theta_{\text{old}}}(o_i|q)} A_i, \\
\operatorname{clip}\left( \frac{\pi_\theta(o_i|q)}{\pi_{\theta_{\text{old}}}(o_i|q)}, 1-\epsilon, 1+\epsilon \right) A_i \Bigg) \Bigg]
\end{split}
\end{equation}
where $\epsilon$ is a positive coefficient that limits the policy updating degree.
\section{Experiment}
\label{sec:experiment}

\subsection{Experiment setting}
\textbf{Implementation Details.}
1) During the SFT training phase, Qwen-2.5-VL-7B-Instruct is adopted as the base model. The training is conducted for 3 epochs. For parameter-efficient fine-tuning, LoRA~\cite{hulora} is employed, with a rank of 64, an alpha of 128, and an initial learning rate of 1e-4.
2) During the RL training phase, the rollout batch size is set to 32, with 8 trajectories sampled per query.
3) In the evaluation phase, Across all benchmarks, the generation temperature is uniformly set to 0.01. Since the base model has not undergone intermediate reasoning training, it is prompted to generate direct answers. Additionally, human evaluation is conducted on QVS-Bench: 100 questions are randomly selected from this benchmark, and 6 human annotators are invited to perform the evaluation to verify the rationality and accuracy of the model’s outputs.

\begin{table*}[t!]
    \centering
    \caption{Results on spatial reasoning benchmarks.  Bold numbers indicate the best performance. For EmbSpatial-Bench, depth is a subset of the far and close sub-tasks. Notations: Cam refers to Camera-based, Persp refers to Perspective Taking, Single refers to Single-view, and Multi refers to Multi-view.}
    \resizebox{\textwidth}{!}{
        \footnotesize
        \setlength{\tabcolsep}{1.5pt} 
        \begin{tabular}{l|cc|ccc|cc|ccc|c}
        \hline
        & \multicolumn{2}{c|}{\textbf{EmbSpatial}} & 
          \multicolumn{3}{c|}{\textbf{ViewSpatial}} & 
          \multicolumn{2}{c|}{\textbf{OmniSpatial}} & 
          \multicolumn{3}{c|}{\textbf{SPAR}} & 
          \multirow{2}{*}{\textbf{Overall}} \\
        & Overall & Depth & Overall & Cam & Person & Overall & Persp & Overall & Single & Multi & \\
        \hline
        \rowcolor{lightgray}\multicolumn{12}{l}{\textit{Closed-source Models}} \\
        GPT-4o & 65.85 & 61.86 & 34.95 & - & - & 47.81 & - & 36.39 & \textbf{38.10} & 35.30 & 46.25 \\
        Gemini 2.0 flash & 62.12 & 61.61 & 32.56 & - & - & \textbf{48.40} & - & 30.08 & - & - & 43.29 \\ 
        \hline
        \rowcolor{lightgray}\multicolumn{12}{l}{\textit{Open-source Models}} \\
        LLaVA-OneVision-7B & \textbf{72.94} & \textbf{57.96} & 27.49 & 28.49 & 26.54 & 35.68 & 38.35 & 31.20 & 33.13 & 29.92 & 41.83 \\ 
        InternVL2.5-8B & 66.21 & 54.06 & 38.13 & 36.80 & \textbf{39.38} & 42.68 & 41.14 & 36.28 & 35.96 & 36.46 & 45.83 \\ 
        \hline
        \rowcolor{lightgray}\multicolumn{12}{l}{\textit{Qwen2.5-VL-7B Based Spatial Models}} \\
        Qwen2.5VL-7B & 66.14 & 52.23 & 37.11 & 39.54 & 34.83 & 40.68 & 41.68 & 32.73 & 32.57 & 28.98 & 44.17 \\ 
        SpaceR-7B & 68.09 & 53.23 & 38.38 & 41.10 & 35.81 & 42.31 & 41.35 & \textbf{37.30} & 37.91 & \textbf{36.82} & 46.52 \\ 
        V2LO-7B (w/o RL) & 67.61 & 54.31 & 38.13 & 39.13 & 36.83 & 42.60 & 45.28 & 34.66 & 33.51 & 35.41 & 45.75 \\ 
        V2LO-7B (Ours) & 68.63 & 56.72 & \textbf{40.18} & \textbf{41.66} & 38.78 & 44.36 & \textbf{46.70} & 36.68 & 36.63 & 36.72 & \textbf{47.46} \\ 
        \hline
        \end{tabular}
    }
    \label{Tab:open_bench}
\end{table*}

\begin{table*}[t] 
    \centering
    \caption{Results on QVS-Bench. The evaluation is categorized by task type. Bold numbers indicate the best performance for each metric (excluding Human baseline).}
    \resizebox{\textwidth}{!}{ 
        \begin{tabular}{l|c|c|c|c|c|c}
        \toprule
        &
        \multicolumn{1}{c|}{\textbf{Relative Distance}} & 
        \multicolumn{1}{c|}{\textbf{Object Count}} & 
        \multicolumn{1}{c|}{\textbf{Min Distance}} & 
        \multicolumn{1}{c|}{\textbf{Vertical Direction}} & 
        \multicolumn{1}{c|}{\textbf{Horizontal Direction}} & 
        \multirow{1}{*}{\textbf{Overall}} \\
        \midrule
        \rowcolor{lightgray}\multicolumn{7}{l}{\textit{Human}} \\
        Human & 74.00 & 56.12 & 59.20 & 62.00 & 63.00 & 62.96 \\ 
        \midrule
        \rowcolor{lightgray}\multicolumn{7}{l}{\textit{Closed-source Models}} \\
        GPT-4o & 57.00 & 42.50 & 24.51 & 22.00 & 24.00 & 36.75 \\ 
        GPT-5 & 68.00 & 46.52 & 28.74 & 36.00 & 26.00 & 43.57 \\
        gemini 2.0 flash & 61.00 & 50.11 & 23.40 & 27.00 & 26.00 & 40.25 \\ 
        gemini 2.5 flash & \textbf{71.00} & 37.40 & 19.65 & 35.00 & 29.00 & 40.01 \\ 
        \midrule
        \rowcolor{lightgray}\multicolumn{7}{l}{\textit{Open-source Models}} \\
        LLaVA-OneVision-7B & 43.00 & 43.12 & 18.80 & 23.00 & 26.00 & 32.36 \\ 
        InternVL2.5-8B & 49.00 & 40.85 & 23.40 & 24.00 & 37.00 & 35.94 \\ 
        \midrule
        \rowcolor{lightgray}\multicolumn{7}{l}
        {\textit{Qwen2.5-VL-7B Based Spatial Models}} \\
        Qwen2.5VL-7B & 45.00 & 46.75 & 14.85 & 25.00 & 27.00 & 33.15 \\ 
        SpaceR-7B & 49.00 & \textbf{62.25} & 24.10 & 23.00 & 27.00 & 40.09 \\ 
        V2LO-7B & 68.00 & 54.20 & \textbf{31.55} & \textbf{73.00} & \textbf{72.00} & \textbf{56.56} \\ 
        \bottomrule
        \end{tabular}
    }
    \label{Tab:task_my_bench}
\end{table*}

\textbf{Benchmarks.}
To comprehensively evaluate the model's spatial reasoning capabilities, we employ a diverse suite of benchmarks. Our evaluation centers on QVS-Bench, supplemented by extensive evaluations of four other well-established benchmarks: EmbSpatial-Bench\cite{du2024embspatial}, ViewSpatial-Bench\cite{li2025viewspatialbenchevaluatingmultiperspectivespatial}, OmniSpatial-Bench\cite{omnispatial25}, and SPAR-Bench\cite{zhang2025flatland}.

\subsection{Main Results}

\textbf{Overall Analysis.}
Detailed results on the four open-source spatial reasoning benchmarks are summarized in \cref{Tab:open_bench}, and results on QVS-Bench are presented in \cref{Tab:task_my_bench}. Based on these experimental findings, we provide key observations and in-depth analyses below.

First, compared to the base model Qwen2.5-VL-7B, V2LO-7B achieves a comprehensive average accuracy of 47.46\% across open benchmarks, representing a 3.29\% absolute improvement. Furthermore, V2LO-7B surpasses the closed-source GPT-4o (46.25\%) and several prominent open-source counterparts, validating the efficacy of our proposed methodology.

Second, V2LO-7B demonstrates robust performance on critical spatial reasoning sub-tasks, yielding 46.70\% accuracy in Perspective Taking (OmniSpatial) and 41.66\% in Camera-based reasoning (ViewSpatial). By outperforming SpaceR-7B in these core dimensions, V2LO-7B underscores that fine-grained coordinate modeling and symbolic mathematical reasoning can effectively augment a model's capacity in spatial reasoning.

Third, evaluations on QVS-Bench, V2LO-7B (56.56\%) outperforms mainstream models including GPT-5 (43.57\%) and the spatially enhanced SpaceR-7B (40.09\%), yet still lags behind the human baseline (62.96\%). This performance gap is mainly concentrated in numerical estimation tasks, as the model fails to match human intuition and accuracy in predicting precise metrics.

Fourth, regarding specific sub-tasks, V2LO-7B presents a competitive advantage in directional judgment, scoring 73.0\% and 72.0\% for vertical and horizontal orientation, respectively. This result not only far exceeds existing models but also surpasses human-level performance on this task. As illustrated in \cref{fig:dataset_example,fig:benchmark}, such superiority arises from our tailored directional task design: it requires the construction of a new local coordinate system in accordance with the specified direction to solve the problem, which demands a strong spatial imagination. Our framework effectively translates abstract qualitative spatial cues into deterministic geometric calculations, thus addressing the complexities of multi-object relative positioning.

\subsection{Analysis Study}

\begin{figure*}[t!]
    \centering
    \begin{subfigure}{0.48\linewidth}
        \centering
        \includegraphics[width=\linewidth]{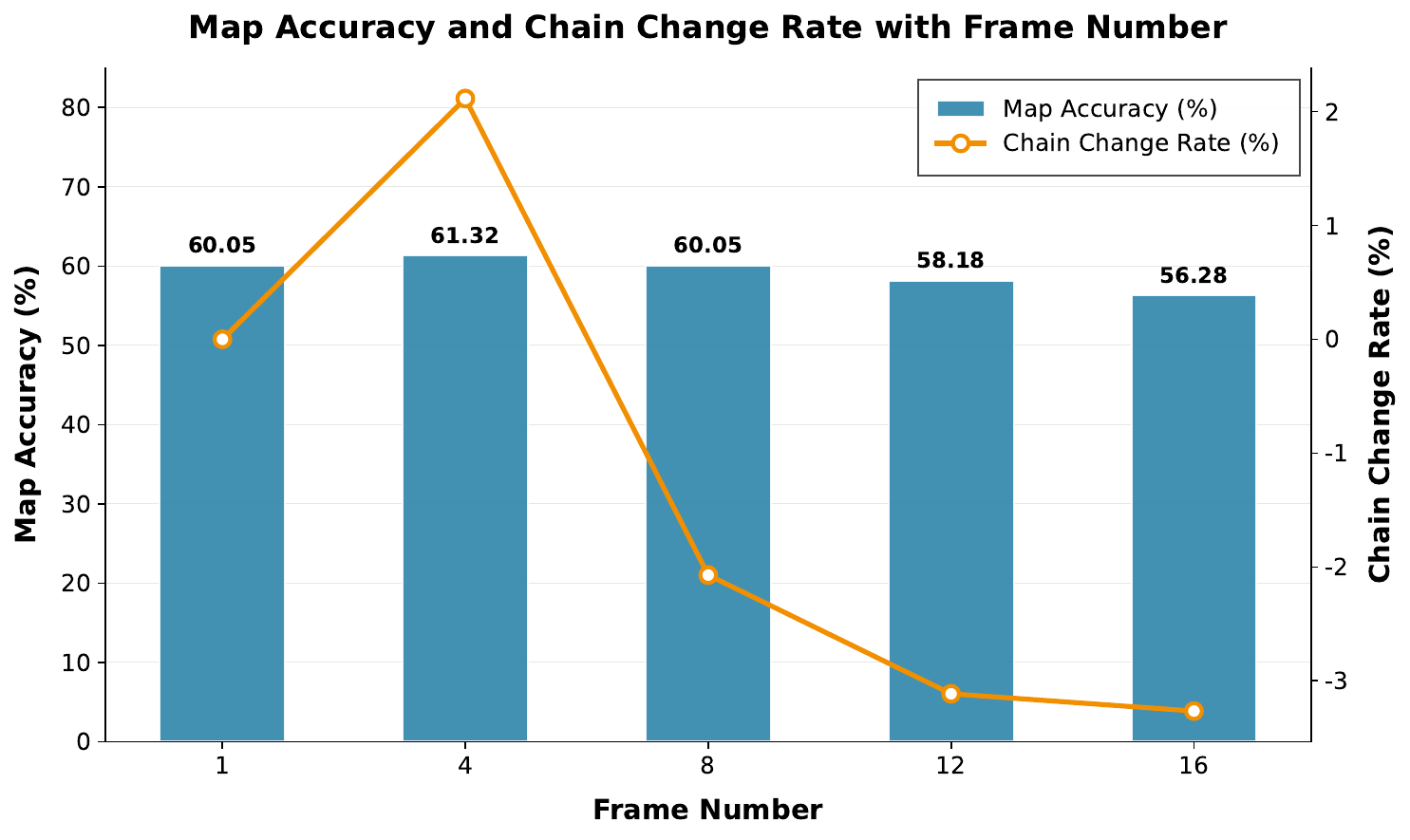}
        \caption{\justifying Impact of the number of input frames on cognitive map accuracy.}
        \label{fig:frame}
    \end{subfigure}
    \hfill
    \begin{subfigure}{0.48\linewidth}
        \centering
        \includegraphics[width=\linewidth]{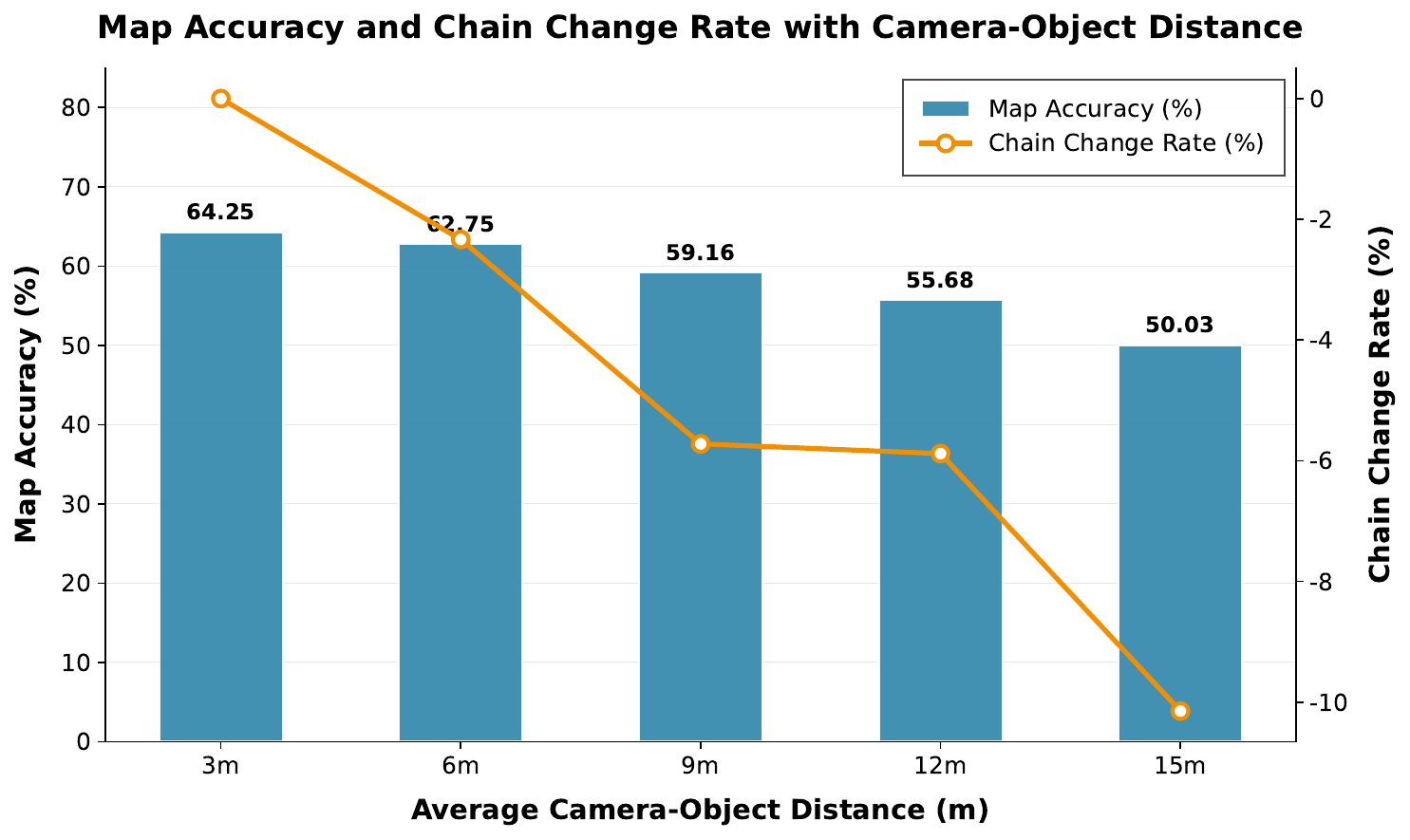}
        \caption{\justifying Impact of object-to-camera distance on cognitive map accuracy.}
        \label{fig:distance}
    \end{subfigure}
    
    \vspace{5mm}
    
    \begin{subfigure}{0.48\linewidth}
        \centering
        \includegraphics[width=\linewidth]{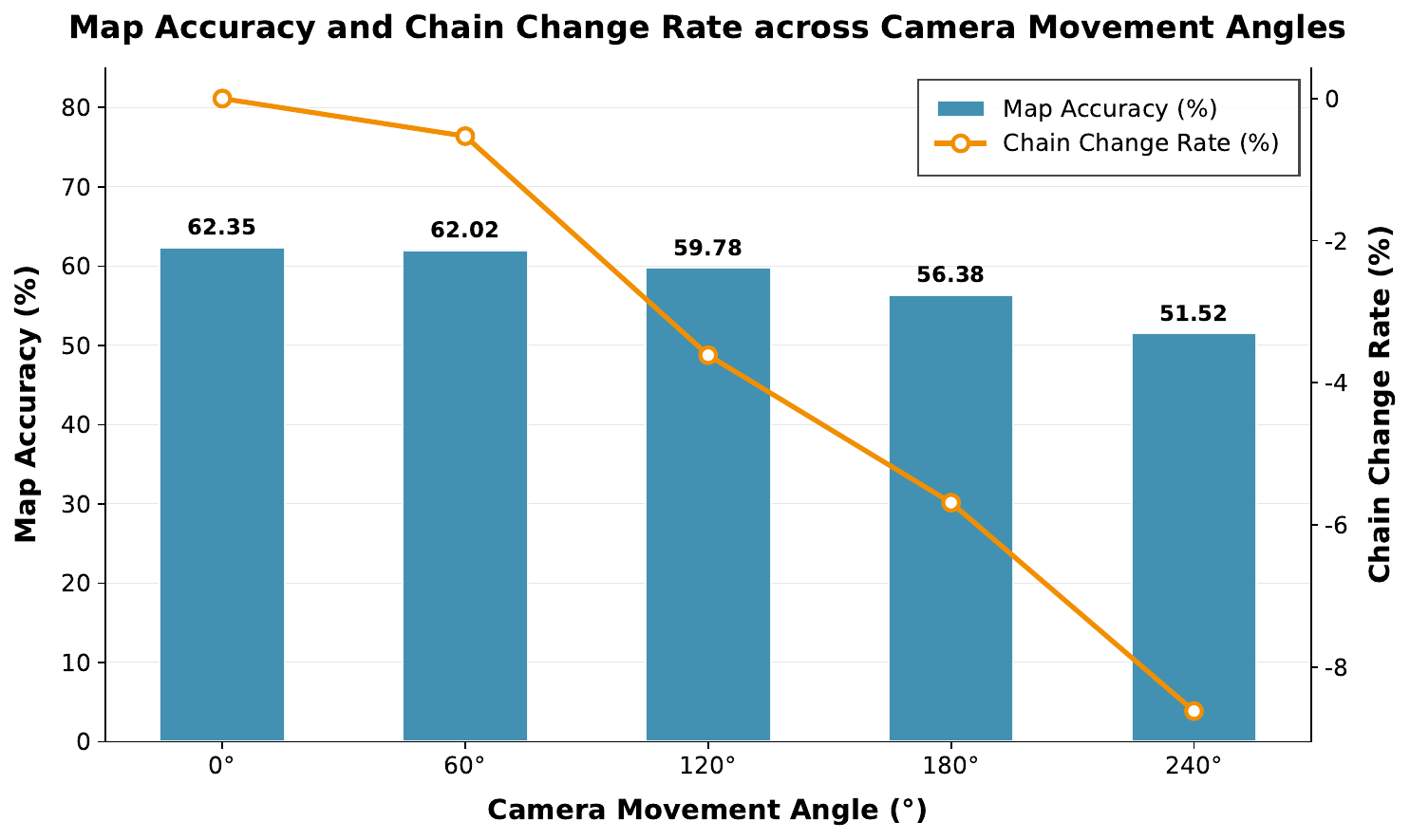}
        \caption{\justifying Impact of cumulative camera rotation on cognitive map accuracy.}
        \label{fig:angle}
    \end{subfigure}
    \hfill
    \begin{subfigure}{0.48\linewidth}
        \centering
        \includegraphics[width=\linewidth]{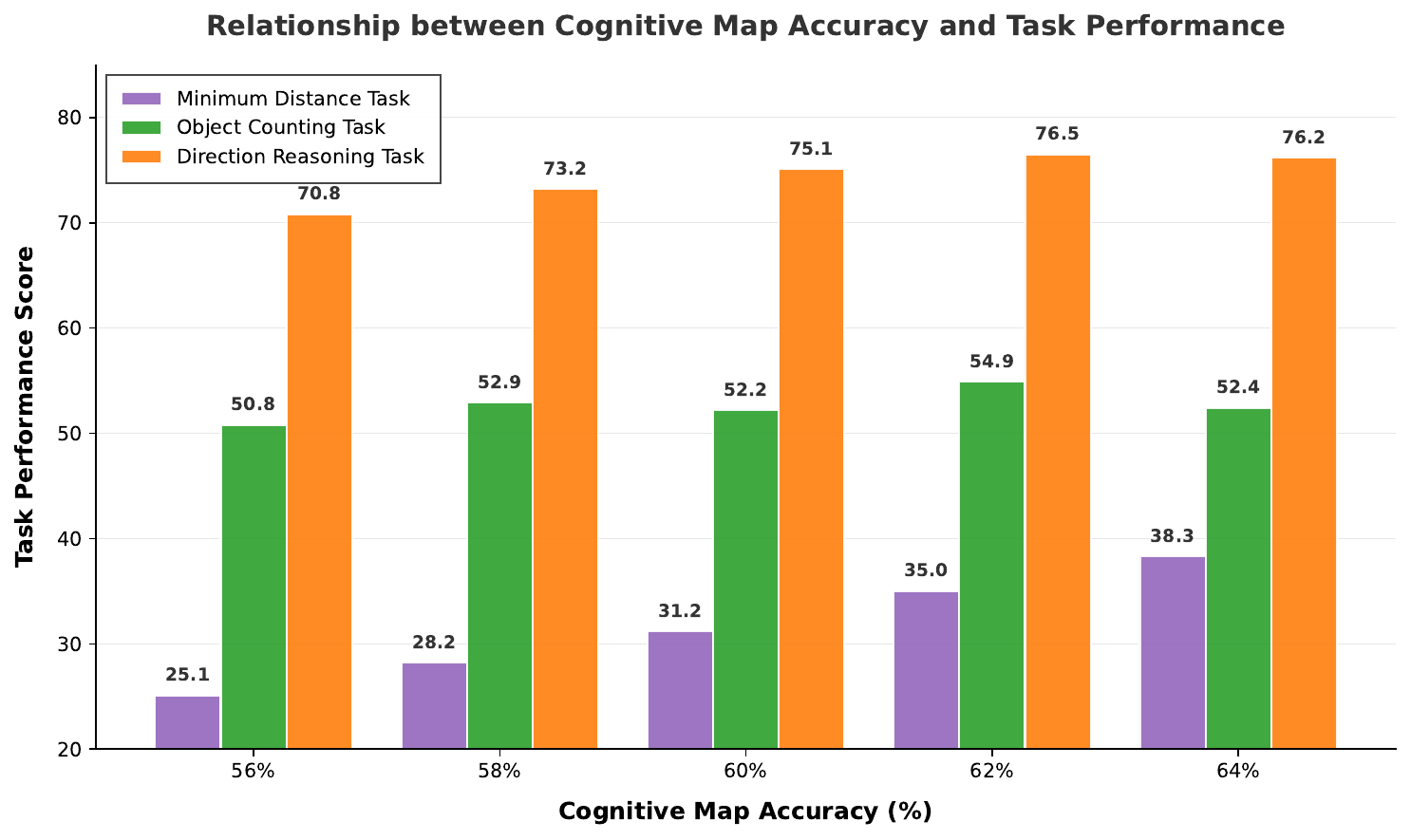}
        \caption{\justifying Correlation between cognitive map accuracy and task performance.}
        \label{fig:map_with_task}
    \end{subfigure}

    \caption{Evaluation of cognitive map accuracy across observational factors and downstream tasks.
    (a)-(c) Cognitive map precision under varying frame counts, object distances, and cumulative rotation angles. 
    (d) Performance of Minimum Distance, Object Counting, and Direction Reasoning tasks at different cognitive map accuracy levels.}
    \label{fig:combined_figures}
\end{figure*}

\textbf{Factors Influencing Cognitive Map Accuracy.}
Our approach is built upon a metric-grounded cognitive map, whose localization and geometric precision constitute a core prerequisite for high-quality downstream spatial reasoning. To rigorously quantify the divergence between the predicted cognitive map and ground-truth layouts, we adopt a simulated dataset that was isolated from the training data, enabling unbiased evaluation of map reconstruction quality.

Considering that many tasks only require correct relative spatial positioning, we refrain from adopting absolute position metrics such as IOU and instead evaluate predictions from a relative perspective. We employ three metrics for accuracy assessment, and use their arithmetic mean as the comprehensive overall score. Specifically:
(1) Bounding box area score is computed by normalizing the area divergence between predicted and ground-truth bounding boxes;
(2) Center-point distance score is obtained by normalizing the relative distance discrepancy between predicted and ground-truth object centroid pairs;
(3) Angular alignment score is generated by normalizing the absolute angle deviation between predicted and ground-truth object layout angles.
The specific calculation formula is as follows~\cref{eq:eval_metrics}.
\begin{equation}
\label{eq:eval_metrics}
S_i =
\begin{cases}
1 - \dfrac{|A_{\text{pred}} - A_{\text{gt}}|}{A_{\text{gt}}}, & i = \text{area} \\
1 - \dfrac{|d(\boldsymbol{c}_{\text{pred}}, \boldsymbol{c}_{\text{gt}})|}{d_{\text{gt}}}, & i = \text{distance} \\
1 - \dfrac{|\theta_{\text{pred}} - \theta_{\text{gt}}|}{180^\circ}, & i = \text{angle}
\end{cases}
\end{equation}
where $A_{\text{pred}}/A_{\text{gt}}$ denote the predicted/ground-truth bounding box area, $\boldsymbol{c}_{\text{pred}}/\boldsymbol{c}_{\text{gt}}$ represent the predicted/ground-truth object centroid coordinate vectors, $\theta_{\text{pred}}/\theta_{\text{gt}}$ are the predicted/ground-truth object layout angles (unit: degree), and $d_{\text{gt}}$ is the ground-truth distance between object centroid pairs (normalization term).

We conduct ablation analyses to dissect performance variations driven by three pivotal factors: (1) the count of input video frames, (2) the mean object-to-camera distance during the video, and (3) the cumulative camera rotation angle throughout the sequence.

As shown in \cref{fig:frame}, the number of input frames exerts a modest influence on accuracy. Performance remains relatively stable in shorter sequences, varying slightly from 60.05\% at 1 frame to a peak of 61.32\% at 4 frames. This suggests that while a few additional views provide detectable gains in spatial reasoning, the benefits are limited. However, as the sequence extends to 16 frames, accuracy declines to 56.28\%, likely due to the accumulation of estimation noise or redundant information that clutters the cognitive map.

In contrast, the average object-to-camera distance significantly impacts map precision. The highest accuracy of 64.25\% is achieved at a close proximity of 3m, but this figure drops to 50.03\% as the distance extends to 15m. This degradation underscores the inherent difficulty of metric estimation for distant objects, where reduced parallax and sparse visual features hinder the accurate inference of dimensions and spatial coordinates.

Cumulative camera rotation also shows a clear inverse correlation with accuracy. As the total rotation increases from 0° to 240°, performance steadily declines from 62.35\% to 51.52\%. Large angular changes pose a challenge to maintaining spatial consistency across shifting viewpoints, as drastic maneuvers likely introduce cumulative estimation drift.

In summary, our analysis reveals that while cognitive map precision is relatively robust to variations in frame count within short sequences, it is notably more sensitive to increased distances and large rotations. These findings highlight the trade-off between visual information gain and the susceptibility to noise in complex spatial reasoning tasks.

\textbf{The Impact of Cognitive Map Accuracy on Task Performance.}
To evaluate the functional utility of the cognitive map, we examine how map precision propagates to downstream task performance. By sampling performance across a spectrum of map accuracy levels from 56\% to 64\%, we observe distinct sensitivity patterns among the three evaluated tasks: Minimum Distance Estimation, Object Counting, and the Direction Reasoning Task. Our results reveal that the impact of cognitive map precision is highly task-dependent.

In \cref{fig:map_with_task}, Minimum Distance Estimation is strongly sensitive to cognitive map accuracy. The task score rises consistently from 25.10\% to 38.30\% as map precision improves. This substantial gain highlights that tasks requiring fine-grained spatial metrics are highly sensitive to the quality of the underlying map, as precise coordinates are indispensable for accurate distance calculations.

In contrast, Object Counting is relatively robust to variations in map precision The performance varies marginally, ranging from 50.80\% to 52.40\% across the sampled accuracy range. This suggests that for discrete reasoning tasks such as counting, the map’s ability to represent object presence is more critical than its absolute metric precision.

Direction Reasoning achieves high baseline performance and improves moderately with map accuracy before plateauing. Accuracy increases from 70.80\% to a peak of 76.50\% before stabilizing. This trend can be predominantly attributed to the fact that direction-oriented reasoning primarily relies on the relative spatial orientation relationships between objects. Such reasoning only requires compliance with basic relative orientation constraints, exhibiting a certain degree of tolerance for prediction errors. Consequently, the task exhibits an inherent robustness to minor metric inaccuracies, yielding higher overall scores even at lower map precision levels.

\begin{table*}[!t]
    \centering
    \small  
    \begin{minipage}{0.48\textwidth}
        \centering
        \caption{Ablation results of reasoning forms. (1) Free-style CoT (2) Grid Maps (from 10x10 to 40x40) use block maps. (3) V2LO-7B uses a metric-grounded map.}
\resizebox{\linewidth}{!}{%
    \begin{tabular}{lcccccc}
        \toprule
        & \textbf{QVS} & \textbf{Emb} & \textbf{View} & \textbf{Omni} & \textbf{SPAR} & \textbf{Overall} \\
        \midrule
        Qwen2.5VL-7B    & 33.15 & 66.14 & 37.11 & 40.68 & 32.73 & 41.96 \\
        \midrule
        Free-style CoT & 38.24 & 64.07 & 40.41 & 41.23 & 32.03 & 43.20 \\
        \midrule
        Grid Map(10x10) & 43.51 & 67.36 & 39.32 & 41.62 & 34.51 & 45.26 \\
        Grid Map(20x20) & 46.45 & 65.69 & 39.53 & 42.99 & 35.54 & 46.04 \\
        Grid Map(30x30) & 46.23 & 67.85 & 38.43 & 43.36 & 34.28 & 46.03 \\
        Grid Map(40x40) & 45.17 & 67.58 & 39.95 & 39.99 & 35.43 & 45.62 \\
        \midrule
        V2LO-7B         & 56.56 & 68.63 & 40.18 & 44.36 & 36.68 & 49.28 \\
        \bottomrule
    \end{tabular}
}
\label{tab:Block map vs detailed map}  
    \end{minipage}
    \hfill  
    \begin{minipage}{0.48\textwidth}
        \centering
        \caption{Ablation results on training recipes.
(1) SFT: Only SFT training; 
(2) RL (Real data): Direct RL using real data without SFT; 
(3) SFT + RL (Simulated data): RL using virtual data; 
(4) V2LO-7B: The complete two-stage training (RL using real data).}
\resizebox{\linewidth}{!}{%
    \begin{tabular}{lcccccc}
        \toprule
        \textbf{Model Configuration} & \textbf{QVS} & \textbf{Emb} & \textbf{View} & \textbf{Omni} & \textbf{SPAR} & \textbf{Overall} \\
        \midrule
        Qwen2.5-VL-7B          & 33.15 & 66.14 & 37.11 & 40.68 & 32.73 & 41.96 \\
        \midrule
        \quad + SFT                    & 38.48 & 67.61 & 38.13 & 42.60 & 34.66 & 44.30 \\
        \quad + RL (Real data)         & 39.65 & 67.25 & 38.62 & 40.90 & 33.60 & 44.00 \\
        \quad + SFT + RL (Simulated data)& 41.51 & 60.22 & 36.71 & 42.47 & 35.67 & 43.32 \\
        \midrule
        V2LO-7B                        & 56.56 & 68.63 & 40.18 & 44.36 & 36.68 & 49.28 \\
        \bottomrule
    \end{tabular}
}
\label{tab:training_stage_ablation}  
    \end{minipage}
\end{table*}

\subsection{Ablation Study}

\textbf{Reasoning Forms Ablation.}
To investigate how spatial representations and reasoning structures influence performance, we conduct an ablation study comparing three distinct paradigms: Free-style CoT, Grid Maps, and our proposed metric-grounded map.

First, we examine the impact of structured reasoning. Compared to the vanilla Qwen2.5VL-7B (41.96\%), the introduction of Free-style CoT (43.20\%) yields a little performance gain. However, this unstructured reasoning approach still falls short of any representation that incorporates explicit spatial structures (Grid or Metric maps). This disparity underscores that while CoT logic is beneficial, spatial reasoning specifically requires a structured representation to bridge the gap between visual tokens and geometric relationships.
Additionally, we observe that numerical computation grounded in cognitive maps eliminates the inherent ambiguity of natural language descriptions via standardized coordinate representations, laying a robust foundation for rigorous spatial geometric computation.

Second, we analyze the limitations of grid-based discretization. Following \cite{yang2025thinking}, we evaluate Grid Maps at resolutions ranging from $10 \times 10$ to $40 \times 40$. As shown in \cref{tab:Block map vs detailed map}, performance peaks at a $20 \times 20$ resolution (46.04\%) but plateaus or even declines at higher granularities (45.62\% at $40 \times 40$). This trend suggests a trade-off between discretization precision and learnability: while higher resolutions theoretically reduce quantization errors, they create an overly fragmented state space that complicates the mapping between discrete tokens and physical coordinates, thereby hindering model optimization.

Finally, equipped with a metric-grounded representation, our V2LO-7B outperforms competing methods and achieves 49.28\% accuracy. Our metric-based approach facilitates fine-grained spatial reasoning and yields a consistent additional improvement in overall accuracy across all benchmarks.

\textbf{Training Phase Ablation.}
We quantify the contribution of each training stage with diverse training configurations. The experimental results validate the efficacy of the structured numerical reasoning paradigm inherent to the SFT framework, and further demonstrate that RL training empowers the model to attain robust generalization capabilities in real-world scenarios.

As illustrated in \cref{tab:training_stage_ablation}, ablation studies further confirm that SFT serves as a crucial preliminary step, providing the model with reliable coordinate perception and numerical reasoning. Relative to the Qwen2.5-VL-7B (41.96\% ), standalone SFT training equips the model with precise object coordinate representations and chain-of-thought reasoning logic, boosting the overall performance to 44.30\%. Our full model V2LO-7B, achieves the best overall score of 49.28\%, which is higher than the counterpart that directly applies RL on real data without SFT (44.00\%). This contrast firmly corroborates that SFT-induced structured numerical reasoning lays a high-quality optimization foundation for the model, highlighting its indispensable role in the whole training pipeline.

To verify that RL training empowers the model to attain robust generalization capabilities in real-world scenarios, we conduct a comparative analysis of RL training strategies based on simulation data and real-world data under the same SFT training setting. The configuration integrating SFT with simulation-data RL registers an overall score of 43.32\%, while adopting real-world data for RL training after SFT substantially elevates the overall performance to 49.28\%. The two-stage V2LO-7B pipeline yields performance improvements, demonstrating that real-world RL training endows the model with strong real-scenario generalization ability and fully validating the efficacy of the proposed training paradigm.
\section{Conclusion}
\label{sec:conclusion}

In this paper, we proposed Video2Layout, a framework designed to reconstruct metric-grounded spatial layouts from video. By shifting from traditional discretized raster representations to continuous object boundary coordinates, our method empowers MLLMs with quantitative spatial computation capabilities. This approach effectively alleviates the inherent ambiguity of natural language when describing complex spatial relationships. Through a two-stage training process—supervised fine-tuning using high-quality AI2THOR-simulated data and a subsequent reinforcement fine-tuning stage—we enhanced the model’s ability to map visual inputs to precise coordinates and its generalization in real-world scenarios. Furthermore, we investigate the diverse factors that influence spatial reasoning and evaluate the quantitative relationship between cognitive map accuracy and task performance. Extensive experiments on mainstream benchmarks demonstrate that our V2LO-7B model achieves an average improvement of 3.24\% over grid map-based baselines, validating the superiority and robustness of our proposed framework.

{
    \small
    \bibliographystyle{ieeenat_fullname}
    \bibliography{main}
}

\section{Benchmarks Description}
\label{sec:benchmarks}

\begin{itemize}
    \item  SPAR-Bench\cite{zhang2025flatland} is specifically designed to measure the spatial understanding of MLLMs. It contains over 7000 QA pairs covering a spectrum of tasks from basic perception to complex spatial reasoning. The benchmark is further divided into single-view and multi view settings, allowing for comprehensive assessment across varying spatial contexts.
    
    \item EmbSpatial-Bench\cite{du2024embspatial} is a benchmark dataset used to evaluate the spatial understanding ability of large visual language models (LVLMs) in embodied tasks. This dataset contains 3,640 multiple-choice questions, covering 294 object categories and 6 spatial relationships. The data is sourced from figurative 3D scenes such as MP3D, ScanNet, and AI2-THOR. The creation process involves automatically extracting spatial relationships from 3D scenes and generating question-and-answer pairs. EmbSpatial-Bench aims to address the issue of evaluating the spatial understanding capabilities of LVLMs in embodied environments, providing crucial support for the development of embodied AI systems.

    \item ViewSpatial-Bench\cite{li2025viewspatialbenchevaluatingmultiperspectivespatial} is a comprehensive benchmark for multi-view spatial positioning recognition, featuring over 5700 carefully selected samples and five task types, used to evaluate the spatial positioning capabilities of visual language models (VLMs) in 3D environments. The dataset is derived from the validation sets of ScanNet and MS-CoCo. Accurate direction labels are generated through an automated 3D annotation process, providing rich spatial relationship data for the training of VLMs.

    \item OmniSpatial\cite{omnispatial25} is a benchmark dataset for diagnosing the limitations of current visual language models in advanced spatial cognition, covering 50 fine-grained tasks and divided into four dimensions: dynamic reasoning, complex spatial logic, spatial interaction, and perspective transformation. It contains 1.3K samples and 1.5K question-answer pairs.

\end{itemize}

\section{Baselines Description}
\label{sec:baseline}

\begin{itemize}
    \item GPT-4o\cite{hurst2024gpt} is a state-of-the-art MLLM developed by OpenAI, exhibiting strong performance across a variety of vision-language tasks.

    \item Gemini 2.0 Flash, is advanced MLLM from Google’s Gemini family. These models have shown leading performance across several video understanding benchmarks. Gemini 2.0 Flash and Gemini 2.5 Pro, in particular, exhibit enhanced abilities in complex reasoning tasks.

    \item LLaVA-OneVision-7B\cite{li2024llava} represents a strong advancement in open-source multimodal language models (LMMs), combining the Qwen2 language backbone with the SigLIP vision encoder. This integration pushes the performance boundaries of open LMMs, particularly in tasks requiring fine-grained visual understanding.

    \item InternVL2.5-78B\cite{chen2024expanding} is a high-performing open-source MLLM that combines InternViT 6B-448px-V2\_5 as the vision encoder with Qwen2.5-7B-Instruct as the LLM backbone.

    \item Qwen2.5-VL-7B-Instruct\cite{Qwen2.5-VL} is part of the Qwen2.5-VL series, which combines the Qwen2.5 language model with a redesigned Vision Transformer(ViT) architecture for enhanced visual grounding and understanding.

    \item SpaceR-7B\cite{ouyang2025spacer} is a high-performing open-source MLLM that excels in spatial reasoning are effectively trained using cognitive mapping methods.   
\end{itemize}

\section{Case Study}
\label{sec:case_study}

\begin{figure*}[htbp]
    \centering
    \includegraphics[width=\linewidth, trim=0cm 2.7cm 0cm 2.7cm, clip]{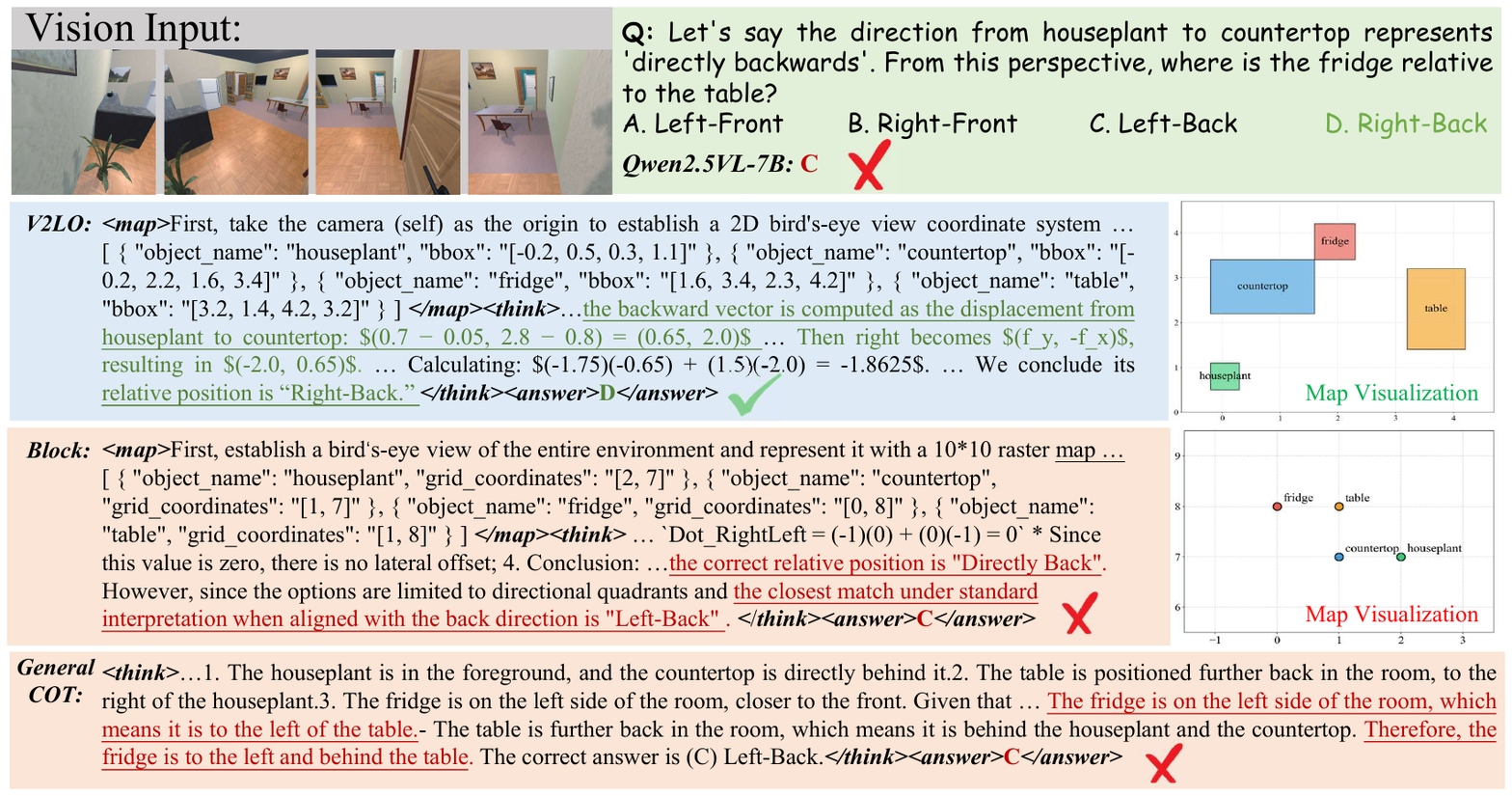}
    \caption{case study (a) in simulation scene.}
    \label{fig:case_study_a}
\end{figure*}

\begin{figure*}[htbp]
    \centering
    \includegraphics[width=\linewidth, trim=0cm 1.7cm 0cm 1.7cm, clip]{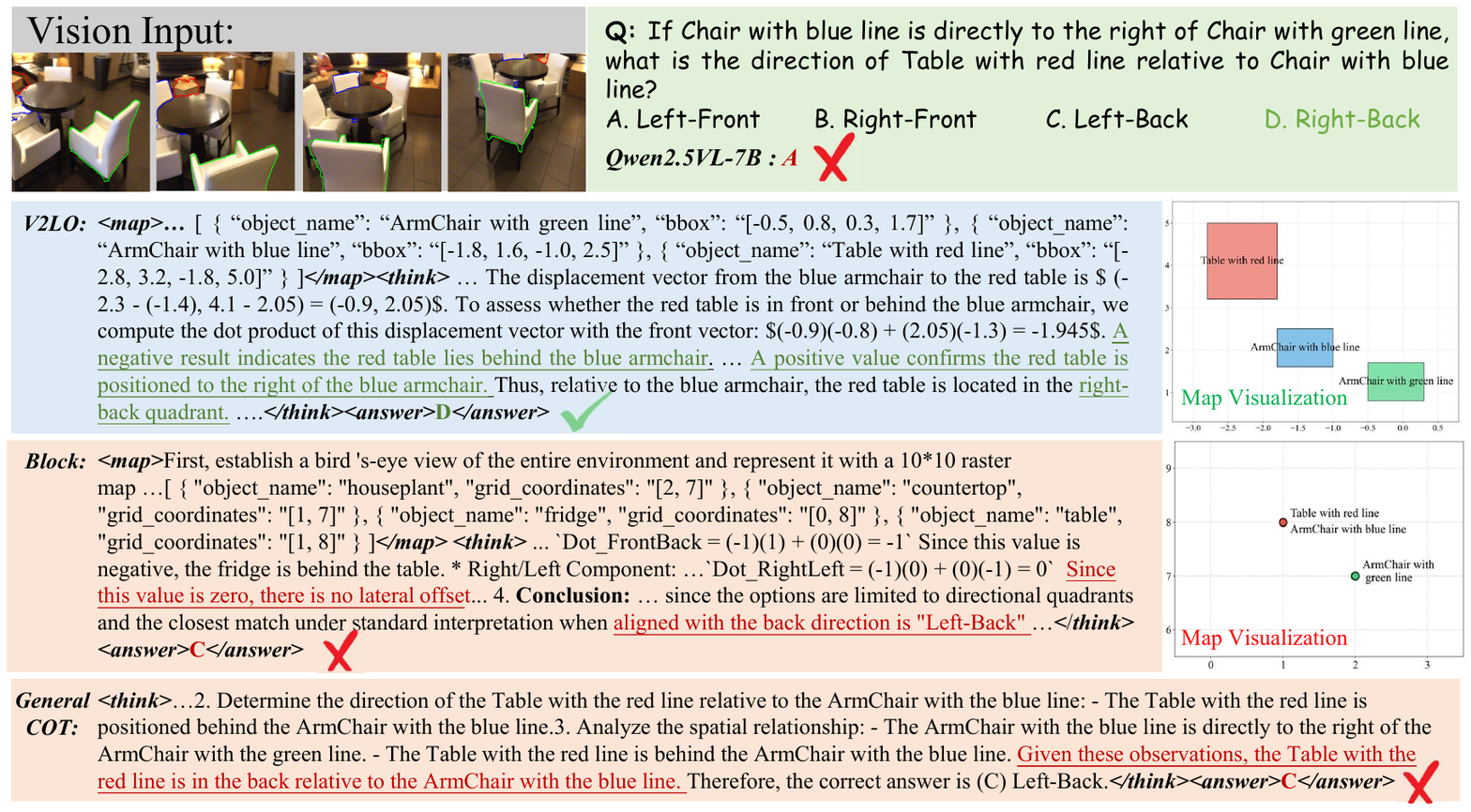}
    \caption{case study (b) in real scene.}
    \label{fig:case_study_b}
\end{figure*}

This study designs two sample validation experiments, conducted in simulated and real physical scenarios.  The performance of the proposed V2LO-7B model is compared with three baselines: Qwen2.5VL-7B-Instruct, the grid map-trained model, and the general COT output model.

In the simulation scenario verification of Case (a) shown in Figure~\ref{fig:case_study_a}, the V2LO-7B model decouples perception and reasoning by means of structured COT. It accurately produces correct results through the deep integration of formal map representation and logical-mathematical reasoning. In contrast, grid maps are unable to distinguish the Y-axis coordinates of the refrigerator and the table, which results in deviations in spatial relationship judgment and ultimately leads to reasoning failure. While SpaceR-7B generates a descriptive narrative, it fails to capture the precise geometric relationships, and Qwen2.5VL-7B fails without a verifiable reasoning process.

In the real-scenario verification of Case (b) shown in Figure~\ref{fig:case_study_b}, the performance advantages of the V2LO-7B model are further validated.  In relatively narrow physical space scenarios, the model can accurately extract and distinguish the coordinate information of each target object.  In contrast, grid map models tend to map different objects into adjacent or even the same grid in compact spatial environments, resulting in the subsequent failure of reasoning tasks. These results underscore the effectiveness and superiority of the refined cognitive map proposed in this study in real and complex scenarios.


\end{document}